\documentclass[11pt,a4paper]{article}
\usepackage{times,latexsym}
\usepackage{url}
\usepackage[T1]{fontenc}
\usepackage[utf8]{inputenc}

\usepackage[acceptedWithA]{tacl2018v2}
\usepackage{amsmath}
\usepackage{multirow}
\usepackage{booktabs}
\usepackage{graphicx}
\usepackage{forest}
\usepackage{qtree}
\usepackage{tikz}
\usetikzlibrary{arrows}

\usepackage{enumitem}

\newcommand{\stackcolor}{red}
\newcommand{\dequecolor}{blue}

\newcommand{\ST}[1]{\textcolor{\stackcolor}{#1}}
\newcommand{\D}[1]{\textcolor{\dequecolor}{#1}}

\usepackage{xspace,mfirstuc,tabulary}

\newif\iftaclinstructions
\taclinstructionsfalse % AUTHORS: do NOT set this to true
\iftaclinstructions
\renewcommand{\confidential}{}
\renewcommand{\anonsubtext}{(No author info supplied here, for consistency with
TACL-submission anonymization requirements)}
\newcommand{\instr}
\fi

\iftaclpubformat % this "if" is set by the choice of options

\else

\fi

\title{Unlexicalized Transition-based\\ 
Discontinuous Constituency Parsing}

\author{
  Maximin Coavoux$^{1}$\Thanks{Work partly done at Université Paris Diderot.}
  \hspace{1em}
  Benoît Crabbé$^{2,3}$
  \hspace{1em}
  Shay B. Cohen$^{1}$ \\
  $^{1}$University of Edinburgh, ILCC\\
  $^{2}$Université Paris Diderot, Université Sorbonne Paris Cité, LLF\\
  $^{3}$Institut Universitaire de France (IUF)\\
  {\tt \{mcoavoux,scohen\}@inf.ed.ac.uk}\\
  {\tt bcrabbe@linguist.univ-paris-diderot.fr}\\
}

\date{}

\begin{document}
\maketitle
\begin{abstract}
    Lexicalized parsing models
    are based on the assumptions that (i) constituents are
    organized around a lexical head (ii) bilexical statistics
    are crucial to solve ambiguities.
    In this paper, we introduce an \textit{unlexicalized}
    transition-based parser for discontinuous constituency structures,
    based on a structure-label transition system
    and a bi-LSTM scoring system.
    We compare it to lexicalized parsing models in order to address the question
    of lexicalization in the context of discontinuous constituency parsing.
    Our experiments show that unlexicalized models systematically
    achieve higher results than lexicalized models,
    and provide additional empirical evidence that
    lexicalization is not necessary to achieve strong parsing results.
    Our best unlexicalized model sets a new state of the art
    on English and German discontinuous constituency treebanks.
    We further provide a per-phenomenon analysis of its errors on discontinuous constituents.
\end{abstract}

\section{Introduction}

\begin{figure}
    \resizebox{0.9\columnwidth}{!}{
        \includegraphics{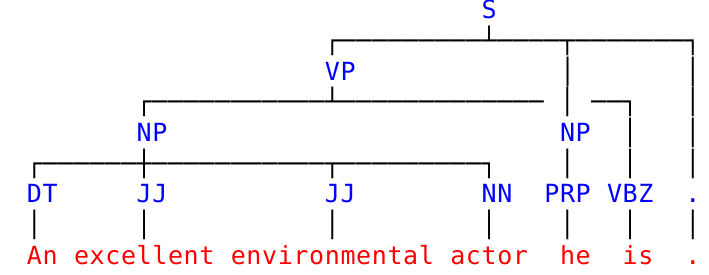}
    }
    \caption{Tree from the Discontinuous Penn Treebank
    \protect\cite{evang-kallmeyer:2011:IWPT}.
    }
    \label{fig:disco-tree}
\end{figure}

This paper introduces an unlexicalized
parsing model and addresses the question of lexicalization,
as a parser design choice,
in the context of transition-based discontinuous constituency parsing.
Discontinuous constituency trees are constituency trees where
crossing arcs are allowed in order to represent long distance
dependencies, and in general phenomena related to word order variations
(e.g.\ the left dislocation in Figure~\ref{fig:disco-tree}).

Lexicalized parsing models \cite{collins:1997:ACL,Charniak:1997:SPC:1867406.1867499}
are based on the assumptions that (i) constituents are
organized around a lexical head (ii) bilexical statistics
are crucial to solve ambiguities.
In a lexicalized PCFG, grammar rules involve
nonterminals annotated with a terminal element that represents
their lexical head, for example:
\[ \text{VP[saw]} \longrightarrow \text{VP[saw]} \; \; \text{PP[telescope]}. \]
The probability of such a rule models the likelihood
that \textit{telescope} is a suitable modifier for \textit{saw}.

In contrast, unlexicalized parsing models renounce
modelling bilexical statistics, based on the assumptions
that they are too sparse to be estimated reliably.
Indeed, \newcite{gildea:01a} observed that removing
bilexical statistics from Collins' \shortcite{collins:1997:ACL}
model lead to at most a 0.5 drop in F-score.
Furthermore, \newcite{bikel:2004:EMNLP} showed that bilexical statistics were in fact
rarely used during decoding, and that when used, they were
close to that of back-off distributions used for unknown word pairs.

Instead, unlexicalized models may rely on grammar rule refinements
to alleviate the strong independence assumptions of PCFGs
\cite{klein-manning:2003:ACL,matsuzaki-miyao-tsujii:2005:ACL,petrov-EtAl:2006:COLACL,narayan-cohen:2016:P16-1}.
They sometimes rely on structural information, such as the boundaries of constituents
\cite{hall-durrett-klein:2014:P14-1,durrett-klein:2015:ACL-IJCNLP,cross-huang:2016:EMNLP2016,stern-andreas-klein:2017:Long,kitaev-klein:2018:Long}.

While initially coined for chart parsers, the notion of lexicalization
naturally transfers to transition-based parsers.
We take \textit{lexicalized} to denote
a model that (i) assigns a lexical head to each constituent (ii) uses
heads of constituents as features to score parsing actions.
Head assignment is typically performed with \textsc{reduce-right} and \textsc{reduce-left}
actions.
Most proposals in transition-based constituency parsing since \newcite{sagae-lavie:2005:IWPT}
have used a lexicalized transition system,
and features involving heads to score actions \cite[among others]{zhu-EtAl:2013:ACL20131,zhang-clark-2011,zhang-clark:2009:IWPT09,crabbe:2014:Coling,wang-mi-xue:2015:ACL-IJCNLP},
including proposals for discontinuous constituency parsing
\cite{versley:2014:SPMRL-SANCL,maier:2015:ACL-IJCNLP,coavoux-crabbe:2017:EACLlong}.
A few recent proposals use an unlexicalized model
\cite{watanabe-sumita:2015:ACL-IJCNLP,cross-huang:2016:EMNLP2016,dyer-EtAl:2016:N16-1}.
Interestingly, these latter models all use recurrent neural networks (RNN)
to compute constituent representations.

Our contributions are the following.
We introduce an unlexicalized discontinuous parsing model,
as well as its lexicalized counterpart.
We evaluate them in identical experimental conditions.
Our main finding is that, in our experiments,
unlexicalized models consistently outperform lexicalized models.
We assess the robustness of this result by performing the comparison
of unlexicalized and lexicalized models with a second pair of transition systems.
We further analyse the empirical properties of the systems in order
to better understand
the reasons for this performance difference.
We find that the unlexicalized system oracle produces shorter, more incremental derivations.
Finally, we provide a per-phenomenon error analysis of our best model
and identify which types of discontinuous constituents are hard to predict.

\newcommand{\smallspace}{\rule{0pt}{5ex}}

\newcommand{\spa}{\;\;}

    \begin{table*}[t]
        \begin{center}
        \begin{tabular}{lrl}
                \toprule
                Structural actions & Input & Output \\
                \midrule
                \textsc{shift}   & $\left\langle S, \spa D, \spa i, \spa  C \right\rangle $ & $ \Rightarrow \left\langle S|D, \spa  \{ i+1 \}, \spa  i+1,\spa  C \right\rangle $\\
                \textsc{merge}   & $\left\langle S|I_{s_0}, \spa D|I_{d_0}, \spa i, \spa  C \right\rangle  $ & $ \Rightarrow  \left\langle S|D, \spa  I_{s_0} \cup I_{d_0}, \spa  i, \spa  C \right\rangle $\\
                \textsc{gap}     & $\left\langle S|I_{s_0}, \spa D,\spa  i,\spa  C \right\rangle $ & $  \Rightarrow  \left\langle S,\spa  I_{s_0}|D, \spa i,\spa  C \right\rangle $\\
                \midrule
                Labelling actions & Input & Output \\
                \midrule
                \textsc{label-X}  & $\left\langle S, \spa  I_{d_0}, \spa i, \spa C \right\rangle $ & $ \Rightarrow  \left\langle S, \spa I_{d_0}, \spa i,\spa  C \cup \{ (X, I_{d_0})\} \right\rangle $\\
                \textsc{no-label} & $\left\langle S, \spa I_{d_0}, \spa i, \spa C \right\rangle $ & $ \Rightarrow  \left\langle S, \spa I_{d_0}, \spa i, \spa  C \right\rangle $\\
                \bottomrule
        \end{tabular}
        \end{center}
        \caption{The
        \textsc{ml-gap} transition system, an unlexicalized transition system for
        discontinuous constituency parsing.}
        \label{unlex:tab:ml-gap}
    \end{table*}

    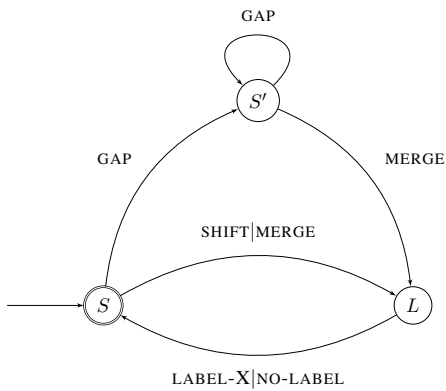
\begin{figure}
        \begin{center}
            \resizebox{0.8\columnwidth}{!}{
                \begin{tikzpicture}
                    \tikzset{vertex/.style = {shape=circle,draw,minimum size=1.5em}}
                    \tikzset{edge/.style = {->,> = latex'}}
                    \node (i) at (-2,0) {};
                    \node[vertex,double] (a) at  (0,0) {$S$};
                    \node[vertex] (b) at  (6,0) {$L$};
                    \node[vertex] (c) at  (3,4) {$S'$};
                    \draw[edge] (i) to (a);
                    \draw[edge] (a) to[bend left] node[label=above:\textsc{shift}$|$\textsc{merge}] {} (b);
                    \draw[edge] (a) to[bend left] node[label=above left:\textsc{gap}] {} (c);
                    \draw[edge] (c) to[loop] node[label=above:\textsc{gap}] {} (c);
                    \draw[edge] (c) to[bend left] node[label=above right:\textsc{merge}] {} (b);
                    \draw[edge] (b) to[bend left] node[label=below:\textsc{label-X}$|$\textsc{no-label}] {} (a);
                \end{tikzpicture}
            }
        \end{center}
        \caption{Action sequences allowed in \textsc{ml-gap}.
        Any derivation must be recognized by the automaton.}
        \label{unlex:fig:action-automata}
    \end{figure}

    \begin{table*}

        \begin{center}
            \begin{tabular}{rll}
                \toprule
                    Structural action & \ST{Stack} -- \D{Dequeue} -- Buffer  & Labelling action \\
                \midrule
                    & An excellent environment actor he is            &               \\
                \textsc{sh}     $\Rightarrow$    & \D{\{An\}$_{d_0}$} excellent environment actor he is     & $\Rightarrow$ \textsc{no-label}  $\Rightarrow$    \\
                \textsc{sh}     $\Rightarrow$    & \ST{\{An\}$_{s_0}$} \D{\{excellent\}$_{d_0}$} environment actor he is   & $\Rightarrow$ \textsc{no-label} $\Rightarrow$ \\
                \textsc{merge}  $\Rightarrow$    & \D{\{An excellent\}$_{d_0}$} environment actor he is   & $\Rightarrow$ \textsc{no-label} $\Rightarrow$          \\
                \textsc{sh}     $\Rightarrow$    & \ST{\{An excellent\}$_{s_0}$} \D{\{environment\}$_{d_0}$} actor he is   & $\Rightarrow$ \textsc{no-label} $\Rightarrow$ \\
                \textsc{merge}  $\Rightarrow$    & \D{\{An excellent environment\}$_{d_0}$} actor he is   & $\Rightarrow$ \textsc{no-label} $\Rightarrow$          \\
                \textsc{sh}     $\Rightarrow$    & \ST{\{An excellent environment\}$_{s_0}$} \D{\{actor\}$_{d_0}$} he is   & $\Rightarrow$ \textsc{no-label} $\Rightarrow$ \\
                \textsc{merge}  $\Rightarrow$    & \D{\{An excellent environment actor\}$_{d_0}$} he is   & $\Rightarrow$ \textsc{label-NP} $\Rightarrow$          \\
                \textsc{sh}     $\Rightarrow$    & \ST{\{An excellent environment actor\}$_{s_0}$} \D{\{he\}$_{d_0}$} is   & $\Rightarrow$ \textsc{label-NP} $\Rightarrow$          \\
                \textsc{sh}     $\Rightarrow$    & \ST{\{An excellent environment actor\}$_{s_1}$ \{he\}$_{s_0}$} \D{\{is\}$_{d_0}$}   & $\Rightarrow$ \textsc{no-label} $\Rightarrow$          \\
                \textsc{gap}    $\Rightarrow$    & \ST{\{An excellent environment actor\}$_{s_0}$} \D{\{he\}$_{d_1}$ \{is\}$_{d_0}$}   &     $\Rightarrow$      \\
                \textsc{merge}  $\Rightarrow$    & \ST{\{he\}$_{s_0}$} \D{\{An excellent environment actor is\}$_{d_0}$}   & $\Rightarrow$ \textsc{label-VP} $\Rightarrow$          \\
                \textsc{merge}  $\Rightarrow$    & \D{\{he An excellent environment actor is\}$_{d_0}$}   & $\Rightarrow$ \textsc{label-S}          \\
                \bottomrule
            \end{tabular}
        \end{center}
        \caption{Example derivation for the tree in Figure~\ref{fig:disco-tree}
        with the \textsc{ml-gap} transition system.}
        \label{unlex:tab:derivation-mlgap}
    \end{table*}

\section{Related Work}

Several approaches to discontinuous constituency parsing have been proposed.
\newcite{Hall2008} reduces the problem to non-projective dependency
parsing, via a reversible transformation, a strategy developped by 
\newcite{fernandezgonzalez-martins:2015:ACL-IJCNLP} and \newcite{corro-leroux-lacroix:2017:EMNLP2017}.
Chart parsers are based on probabilistic LCFRS \cite{evang-kallmeyer:2011:IWPT,kallmeyer-maier:2010:PAPERS},
the Data-Oriented Parsing (DOP) framework \cite{vancranenburgh2013disc,vancranenburgh2016disc},
or pseudo-projective parsing \cite{versley:2016:DiscoNLP}.

Some transition-based discontinuous constituency parsers 
use the \textit{swap} action, adapted from dependency parsing \cite{nivre:2009:ACLIJCNLP}
either with an easy-first strategy \cite{versley:2014:SPMRL-SANCL,DBLP:journals/corr/Versley14}
or with a shift-reduce strategy 
\cite{maier:2015:ACL-IJCNLP,maier-lichte:2016:DiscoNLP,stanojevic-garridoalhama:2017:EMNLP2017}.
Nevertheless, the swap strategy tends to produce long derivations (in number of actions)
to construct discontinuous constituents; as a result, the choice of an oracle
that minimizes the number of swap actions has a substantial positive effect
in accuracy \cite{maier-lichte:2016:DiscoNLP,stanojevic-garridoalhama:2017:EMNLP2017}.

In contrast, \newcite{coavoux-crabbe:2017:EACLlong} extended
a shift-reduce transition system to handle discontinuous constituents.
Their system allows binary reductions to apply to
the top element in the stack, and any other element in the stack
(instead of the two top elements in standard shift-reduce parsing).
The second constituent for a reduction is chosen dynamically,
with an action called \textsc{gap} that gives access to older
elements in the stack
and can be performed several times before a reduction.
In practice, they made the following modifications over a standard shift-reduce system:
\begin{enumerate}
    \item the stack, that stores subtrees being constructed,
    is split into two parts $S$ and $D$;
    \item reductions are applied to the respective tops of $S$ and $D$;
    \item the \textsc{gap} action, pops an element from $S$ and adds it to $D$,
    making the next element of $S$ available for a reduction.
\end{enumerate}
Their parser outperforms swap-based systems.
However, they only experiment
with a linear classifier, and assume access to
gold part-of-speech (POS) tags for most of their experiments.

All these proposals use a lexicalized model, as defined in the introduction:
they assign heads to new constituents and use them as features to inform parsing decisions.
Previous work on unlexicalized transition-based parsing models
only focused on projective constituency trees \cite{dyer-EtAl:2016:N16-1,DBLP:journals/corr/LiuZ17aa}.
In particular, \newcite{cross-huang:2016:EMNLP2016}
introduced a system that does not require explicit binarization.
Their system decouples the construction of a tree and the labelling
of its nodes by assigning types (\textit{structure} or \textit{label})
to each action, and alternating between a structural action for even steps
and labelling action for odd steps.
This distinction arguably makes each decision simpler.

\section{Transition Systems for Discontinuous Parsing}
\label{sec:transition-systems}

This section introduces an unlexicalized transition system able
to construct discontinuous constituency trees (Section~\ref{sec:mlgap}),
its lexicalized counterpart (Section~\ref{sec:mlgap-lex}),
and corresponding oracles (Section~\ref{sec:oracle}).

\subsection{The Merge-Label-Gap Transition System}
\label{sec:mlgap}

The Merge-Label-Gap transition system (henceforth \textsc{ml-gap})
combines the distinction between structural and labelling actions
from \newcite{cross-huang:2016:EMNLP2016}
and the \textsc{shift-reduce-gap} (\textsc{sr-gap}) strategy with a split stack
from \newcite{coavoux-crabbe:2017:EACLlong}.

Like the \textsc{sr-gap} transition system, \textsc{ml-gap}
is based on three data structures: a stack $S$,
a double-ended queue (dequeue) $D$ and a buffer $B$.
We define a parsing configuration as a quadruple
$\langle S, D, i, C \rangle$, where $S$ and $D$ are
sequences of index sets, $i$ is the index of the
last shifted token,
and $C$ is a set of \textbf{instantiated discontinuous constituents}.
We adopt a representation of instantiated constituents
as pairs $(X, I)$,
where $X$ is a nonterminal label, and $I$ is a set of token indexes.
For example, the discontinuous VP in Figure~\ref{fig:disco-tree}
is the pair (VP, $\{1, 2, 3, 4, 6\}$), because
it spans tokens 1 through 4 and token 6.

The \textsc{ml-gap} transition system is defined as a deduction system in
Table~\ref{unlex:tab:ml-gap}. The available actions
are the following:
\begin{itemize}[noitemsep,topsep=0pt]
    \item The \textsc{shift} action pushes the singleton $\{ i + 1 \}$
    onto $D$.
    \item The \textsc{merge} action removes $I_{s_0}$ and $I_{d_0}$ from
    the top of $S$ and $D$, computes their union $I = I_{s_0} \cup I_{d_0}$,
    transfers the content of $D$ to $S$ and pushes $I$ onto $D$.
    It is meant to construct incrementally subsets of tokens that are constituents.
    \item The \textsc{gap} action removes the top element from $S$
    and pushes it at the beginning of~$D$.
    This action gives the system the ability to construct discontinuous trees,
    by making older elements in $S$ accessible to a subsequent merge
    operation with $I_{d_0}$.
    \item \textsc{label-X} creates a new constituent labelled $X$ whose
    yield is the set $I_{d_0}$ at the top of $D$.
    \item \textsc{no-label} has no effect.
\end{itemize}

Actions are sucategorized into \textbf{structural actions} (\textsc{shift},
\textsc{merge}, \textsc{gap}) and \textbf{labelling actions} (\textsc{no-label},
\textsc{label-X}).
This distinction is meant to make each single decision easier.
The current state of the parser determines the type of action to be predicted next,
as illustrated by the automaton in Figure~\ref{unlex:fig:action-automata}.
When it predicts a projective tree, the parser alternates structural and
labelling actions (states $S$ and $L$).
However, it must be able to perform several \textsc{gap} actions in a row
to predict some discontinuous constituents.
Since the semantics of the \textsc{gap} action, a structural action,
is not to modify the top of $D$, but to make an index set in $S$ available for a \textsc{merge},
it must not be followed by a labelling action.
Each \textsc{gap} action must be followed by either another \textsc{gap}
or a \textsc{merge} action (state $S'$ in the automaton).
We illustrate the transition system with a full derivation in Table~\ref{unlex:tab:derivation-mlgap}.

\subsection{Lexicalized Transition System}
\label{sec:mlgap-lex}

In order to assess the role of lexicalization in parsing,
we introduce a second transition system, \textsc{ml-gap-lex},
which is designed (i) to be lexicalized (ii) to differ minimally from \textsc{ml-gap}.

We define an instantiated \textbf{lexicalized discontinuous constituent} as a
triple $(X, I, h)$ where $X$ is a nonterminal label, $I$ is the set of terminals
that are in the yield of the constituent, and $h \in I$ is the lexical head
of the constituent.
In \textsc{ml-gap-lex}, the dequeue and the stack contains pairs
$(I, h)$, where $I$ is a set of indices and $h \in I$
is a distinguished element of $I$.

The main difference of \textsc{ml-gap-lex} with \textsc{ml-gap} is that there are two \textsc{merge}
actions, \textsc{merge-left} and \textsc{merge-right}, and that each of them
assigns the head of the new set of indexes (and implicitly creates a new directed dependency arc):

\begin{itemize}[noitemsep]
    \item \textsc{merge-left}:
    \begin{tabular}{l}
        $\left\langle S|(I_{s_0}, h_{s_0}),\spa  D|(I_{d_0}, h_{d_0}), \spa i,\spa  C \right\rangle $ \\
        $\Rightarrow \left\langle S|D,\spa  (I_{s_0} \cup I_{d_0}, h_{s_0}),\spa  i,\spa  C  \right\rangle$;
    \end{tabular}
    \item \textsc{merge-right}:
    \begin{tabular}{l}
        $\left\langle S|(I_{s_0}, h_{s_0}),\spa  D|(I_{d_0}, h_{d_0}), \spa i,\spa  C \right\rangle $\\
        $\Rightarrow \left\langle S|D,\spa  (I_{s_0} \cup I_{d_0}, h_{d_0}), \spa i,\spa  C \} \right\rangle $.
    \end{tabular}
\end{itemize}

\begin{figure*}[t]
    \resizebox{\textwidth}{!}{
        \includegraphics{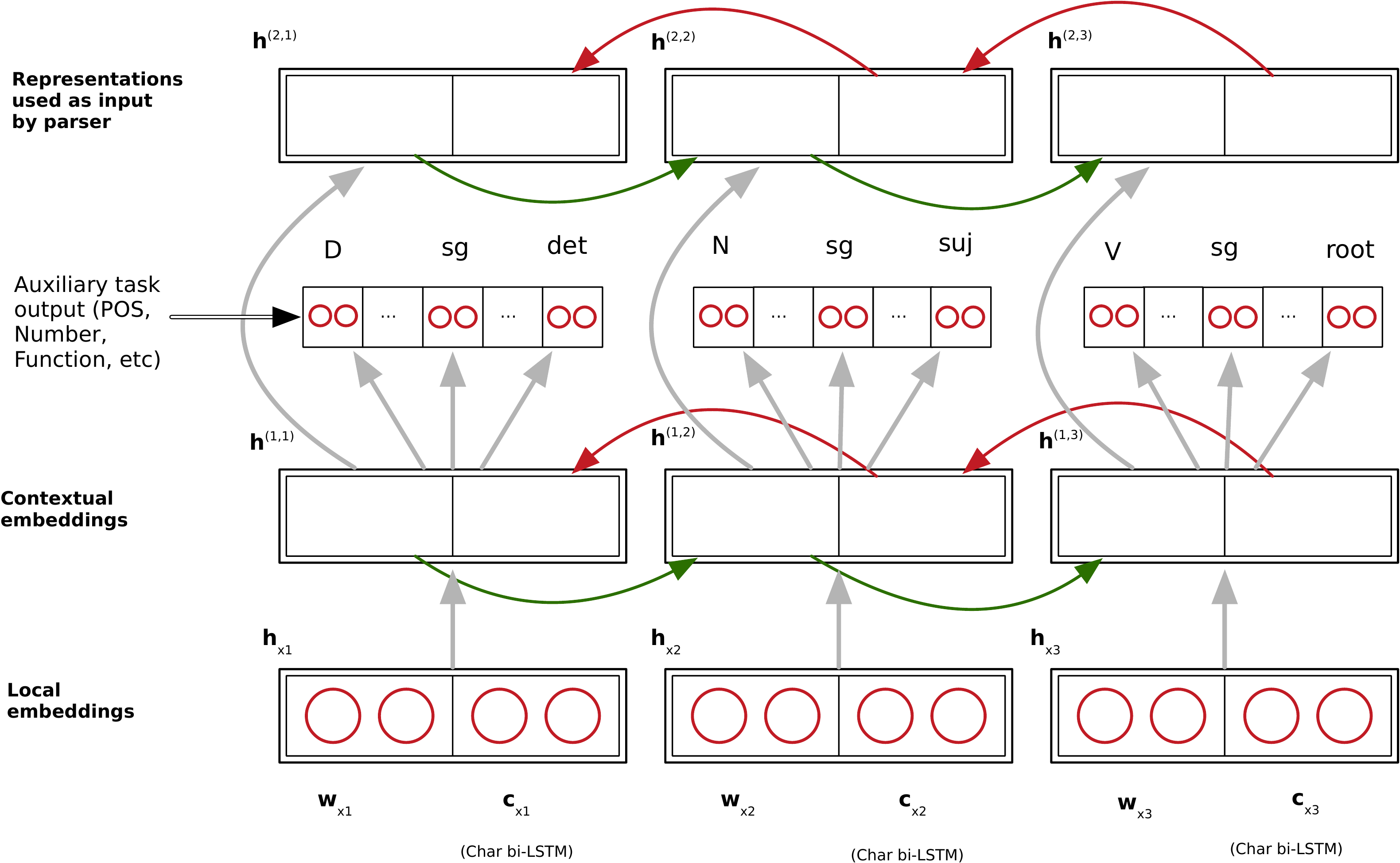}
    }
    \caption{Bi-LSTM part of the neural architecture.
    Each word is represented by the concatenation of
    a standard word-embedding $\mathbf{w}$ and the output
    of a character bi-LSTM $\mathbf{c}$.
    The concatenation is fed to a two-layer bi-LSTM transducer
    that produces contextual word representations.
    The first layer serves as input to the tagger (Section~\ref{sec:tagger}),
    whereas the second layer is used by the parser to instantiate
    feature templates for each parsing step (Section~\ref{sec:parser}).}
    \label{fig:multinet}
\end{figure*}

\subsection{Oracles}
\label{sec:oracle}

In this work, we use deterministic static oracles.
We briefly describe an oracle that builds constituents
from their head outwards (head-driven oracle)
and an oracle that performs merges as soon as possible (eager oracle).
The latter can only be used by an unlexicalized system.

\paragraph{Head-driven oracle}
The head-driven oracle
can be straightforwardly
derived from the oracle for \textsc{sr-gap} presented by \newcite{coavoux-crabbe:2017:EACLlong}.
A derivation in \textsc{ml-gap-lex} can be computed
from a derivation in \textsc{sr-gap} by
(i) replacing \textsc{reduce-left-X} (resp.\ \textsc{reduce-right-X}) actions
by a \textsc{merge-left} (resp.\ \textsc{merge-right}),
(ii) replacing \textsc{reduce-unary-X} actions by \textsc{label-X},
(iii) inserting \textsc{label-X} and \textsc{no-label} actions as required.
This oracle attaches the left dependents of a head first.
In practice, other oracle strategies are possible as long as constituents
are constructed from their head outward.

\paragraph{Eager oracle} For the \textsc{ml-gap} system, we use an oracle
that builds every $n$-ary constituent in a left-to-right fashion,
as illustrated by the derivation in Table~\ref{unlex:tab:derivation-mlgap}.\footnote{The systems exhibit
spurious ambiguity for constructing $n$-ary ($n>2$) constituents.
We leave the exploration of nondeterministic oracles to future work.}
This implicitly corresponds to a left-branching binarization.

\section{Neural Architecture}
\label{sec:architecture}

The statistical model we used is based on a bi-LSTM transducer that
builds context-aware representations for each token in the sentence
\cite{TACL885,cross-huang:2016:P16-2}.
The token representations are then fed as input to
(i) a tagging component for assigning POS tags (ii) a parsing component
for scoring parsing actions.\footnote{A more involved strategy would be to rely on
Recurrent Neural Network Grammars \cite{dyer-EtAl:2016:N16-1,kuncoro-EtAl:2017:EACLlong}.
However, the adaptation of this model to discontinuous parsing
is not straightforward and we leave it to future work.}
The whole architecture is trained end-to-end.
We illustrate the bi-LSTM and the tagging components in Figure~\ref{fig:multinet}.

In the following paragraphs, we describe the architecture that builds shared
representations (Section~\ref{sec:bilstm}), the tagging component (Section~\ref{sec:tagger})
the parsing component (Section~\ref{sec:parser}) and the objective function (Section~\ref{sec:training}).

\begin{table}
    \resizebox{\columnwidth}{!}{
        \begin{tabular}{l l}
            \toprule
            \multicolumn{2}{l}{Configuration:}\\
            \multicolumn{2}{l}{ $\langle S|(I_{s_1}, h_{s_1})|(I_{s_0}, h_{s_0}), D|(I_{d_1}, h_{d_1})|(I_{d_0}, h_{d_0}), i, C \rangle$}\\
            \midrule
            Template set                    & Token indexes       \\
            \midrule
            \textsc{base}       & $\max(I_{s_1}), \min(I_{s_0}), \max(I_{s_0}), \max(I_{d_1}),$\\
                                & $\min(I_{d_0}), \max(I_{d_0}), i$    \\
    \midrule
            \textsc{+lex}       & \textsc{base+} $h_{d_0}, h_{d_1}, h_{s_0}, h_{s_1}$    \\
            \bottomrule
        \end{tabular}
    }
    \caption{Feature template set descriptions.}
    \label{unlex:tab:feature-templates}
\end{table}

\subsection{Building Context-aware Token Representations}
\label{sec:bilstm}

We use a hierarchical bi-LSTM \cite{plank-sogaard-goldberg:2016:P16-2}
to construct context-aware vector representations for each token.
A lexical entry $x$ is represented by the concatenation $\mathbf h_{x}=[\mathbf w_x; \mathbf c_x]$,
where $\mathbf w_x$ is a standard word embedding and $\mathbf c_x = \text{bi-LSTM}(x)$
is the output of a character bi-LSTM encoder, i.e.\ the concatenation
of its last forward and backward states.

We run a sentence-level bi-LSTM transducer over the sequence of local
embeddings $(\mathbf h_{x_1}, \mathbf h_{x_2}, \dots, \mathbf h_{x_n})$,
to obtain vector representations that depend on the whole sentence:
\begin{align*}
    (\mathbf h^{(1)}, \dots, \mathbf h^{(n)})
        = \text{bi-LSTM}(\mathbf h_{x_1}, \mathbf h_{x_2}, \dots, \mathbf h_{x_n}).
\end{align*}
In practice, we use a two-layer bi-LSTM, in order to supervise
parsing and tagging at different layers, following results
by \newcite{sogaard-goldberg:2016:P16-2}.
In what follows, we denote
the $i^{th}$ state of the $j^{th}$ layer with $\mathbf h^{(j,i)}$.

\subsection{Tagger}
\label{sec:tagger}

We use the context-aware representations as input to a softmax
classifier to output a probability distribution over part-of-speech (POS)
tags for each token:
\[ P(t_i = \cdot |x_1^n; \boldsymbol \theta_t) = \text{Softmax}(\mathbf W^{(t)} \cdot \mathbf h^{(1,i)} + \mathbf b^{(t)}), \]
where $\mathbf W^{(t)}, \mathbf b^{(t)} \in \boldsymbol \theta_t$ are parameters.

In addition to predicting POS tags, we also predict other
morphosyntactic attributes when they are available (i.e.\ for the Tiger corpus)
such as the case, tense, mood, person, gender,
since the POS tagset does not necessarily contain this information.
Finally, we predict the syntactic functions of tokens, since this auxiliary
task has been shown to be beneficial
for constituency parsing \cite{coavoux-crabbe:2017:EACLshort}.

For each type of label $l$, we use a separate softmax classifier,
with its own parameters~$\mathbf W^{(l)}$ and~$\mathbf b^{(l)}$:
\[ P(l_i = \cdot |x_1^n; \boldsymbol \theta_t) = \text{Softmax}(\mathbf W^{(l)} \cdot \mathbf h^{(1,i)} + \mathbf b^{\text{(l)}}). \]
For a given token, the number and types of morphosyntactic attributes
depend on its POS tag.
For example, a German noun has a gender and number but no tense nor mood.
We use a default value (`undef') to make sure that every token
has the same number of labels.

\subsection{Parser}
\label{sec:parser}

We decompose the probability of a sequence of actions $a_1^m = (a_1, a_2, \dots, a_m)$
for a sentence $x_1^n$ as the product of probability of individual actions:
\[ P(a_1^m | x_1^n; \boldsymbol \theta_p) = \prod_{i=1}^m P(a_i | a_1^{i-1}, x_1^n;     \boldsymbol \theta_p). \]
The probability of an action given a parsing configuration is computed
with a feed-forward network with two hidden layers:
\begin{align*}
    \mathbf o^{(1)} &= g(\mathbf W^{(1)} \cdot \boldsymbol \Phi_f(a_1^{i-1}, x_1^n) + \mathbf b^{(1)}),  \\
    \mathbf o^{(2)} &= g(\mathbf W^{(2)} \cdot \mathbf o^{(1)} + \mathbf b^{(2)}),\\
    P(a_i | a_1^{i-1}, x_1^n) &= \text{Softmax}(\mathbf W^{(3)} \cdot \mathbf o^{(2)} + \mathbf b^{(3)}),
\end{align*}
where
\begin{itemize}[noitemsep,topsep=0pt]
    \item $g$ is an activation function (rectifier);
    \item $\mathbf W^{(i)}, \mathbf b^{(i)} \in \boldsymbol \theta_p$ are parameters;
    \item $\boldsymbol \Phi_f$ is a function, parameterized by a feature
    template list $f$, that outputs the concatenation of instantiated features,
    for the configuration obtained after performing the sequence of action $a_1^{(i-1)}$
    to the input sentence $x_1^n$.
\end{itemize}

Feature templates describe a list of positions in a configuration.
Features are instantiated by the context-aware representation of the token occupying the position.
For example, token $i$ will yield vector $\mathbf h^{(2,i)}$, the output of the
sentence-level bi-LSTM transducer at position $i$.
If a position contains no token, the feature is instantiated by a special trained embedding.

\paragraph{Feature Templates}
The two feature template sets we used are presented in Table~\ref{unlex:tab:feature-templates}.
The \textsc{base} templates form a minimal set that extracts 7 indexes from
a configuration, relying only on constituent boundaries.
The \textsc{+lex} feature set adds information about the heads of constituents
at the top of $S$ and $D$, and can only be used together with a lexicalized
transition system.

\subsection{Objective Function}
\label{sec:training}

The objective function for a single sentence $x_1^n$ decomposes in a tagging objective
and a parsing objective.
The tagging objective is the negative log-likelihood of gold labels for each token:
\[ \mathcal{L}_t(x_1^n; \boldsymbol \theta_t) = - \sum_{i=1}^n \sum_{j=1}^k  \log P(t_{i,j} | x_1^n; \boldsymbol \theta_t), \]
where $k$ is the number of types of labels to predict.

The parsing objective is the negative log-likelihood of the gold derivation,
as computed by the oracle:
\[ \mathcal{L}_p(x_1^n; \boldsymbol \theta_p) = - \sum_{i=1}^m \log P(a_i | a_1^{i-1}, x_1^n; \boldsymbol \theta_p). \]

We train the model by minimizing $\mathcal{L}_t + \mathcal{L}_p$ over the whole corpus.
We do so by repeatedly sampling a sentence, performing one optimization step for $\mathcal{L}_t$
followed by one optimization step for $\mathcal{L}_p$.
Some parameters are shared across the parser and the tagger, namely the word
and character embeddings, the parameters for the character bi-LSTM
and those for the first layer of the sentence-bi-LSTM.

\begin{table*}
    \begin{center}
        \begin{tabular}{lll cc cc cc}
            \toprule
            &&&\multicolumn{2}{c}{English} & \multicolumn{2}{c}{German (Tiger)} & \multicolumn{2}{c}{German (Negra)} \\
            Transition System & Features & Oracle  & F & Disc. F & F & Disc. F & F & Disc. F \\
            \midrule
            \textsc{ml-gap}          & \textsc{base} & eager  & \bf{91.2} & 72.0 & \bf{87.6} & \bf{60.5} & 83.7 & \bf{53.8} \\
            \textsc{ml-gap} & \textsc{base} & head-driven  & 91.1 & \bf{73.7} & 87.2 & 59.7 & 83.7 & \bf{53.8} \\
            \midrule
            \textsc{ml-gap-lex}      & \textsc{base} & head-driven  & 91.1 & 68.2 & 86.5 & 56.3 & 82.4 & 47.0 \\
            \textsc{ml-gap-lex}      & \textsc{+lex} & head-driven  & 90.9 & 68.2 & 86.5 & 57.0 & 82.7 & 52.3 \\
            \midrule
            \textsc{sr-gap-unlex} & \textsc{base} & eager  & 91.0 & 72.9 & 87.1 & \bf{60.5} & \bf{84.1} & 52.0 \\
            \midrule
            \textsc{sr-gap-lex}          & \textsc{base} & head-driven  & 90.8 & 70.3 & 86.0 & 56.2 & 82.1 & 44.9 \\
            \textsc{sr-gap-lex}          & \textsc{+lex} & head-driven  & 90.8 & 71.3 & 86.5 & 55.5 & 82.8 & 50.6 \\
            \bottomrule
        \end{tabular}
    \end{center}
    \caption[Discontinuous parsing results (development).]{Discontinuous parsing
    results on the development sets.}
    \label{unlex:tab:parse-dev}
\end{table*}

\section{Experiments}
\label{sec:experiments}

The experiments we performed aim at assessing the role of lexicalization
in transition-based constituency parsing.
We describe the data (Section~\ref{sec:data}), the optimization protocol
(Section~\ref{sec:hyperparameters}). Then, we discuss
empirical runtime efficiency (Section~\ref{sec:runtime}),
before presenting the results of our
experiments (Section~\ref{sec:results}).

\subsection{Data}
\label{sec:data}

To evaluate our models, we used the Negra corpus \cite{skut-EtAl:1997:ANLP},
the Tiger corpus \cite{BrantsEtAl02} and the discontinuous version
of the Penn Treebank \cite{evang-kallmeyer:2011:IWPT,J93-2004}.

For the Tiger corpus, we use the SPMRL split \cite{seddah-EtAl:2013:SPMRL}. We obtained the dependency labels
and the morphological information for each token from the dependency treebank
versions of the SPMRL release.
We converted the Negra corpus to labelled dependency trees with the \textsc{depsy}
tool\footnote{\url{https://nats-www.informatik.uni-hamburg.de/pub/CDG/DownloadPage/cdg-2006-06-21.tar.gz}
We modified \textsc{depsy} to keep the same tokenization as
the original corpus.} in order to annotate each token with a dependency label.
We do not predict morphological attributes for the Negra corpus (only POS tags)
since only a small section is annotated with a full morphological analysis.
We use the standard split \cite{dubey-keller:2003:ACL} for this corpus, and no limit
on sentence length.
For the Penn Treebank, we use the standard split (sections~2-21 for training, 22~for development and 23~for test).
We retrieved the dependency labels from the dependency version of the PTB,
obtained by the Stanford Parser \cite{Marneffe06generatingtyped}.

We used the relevant module of discodop\footnote{\url{https://github.com/andreasvc/disco-dop}}
\cite{vancranenburgh2016disc} for evaluation.
It provides an F1 measure on labelled constituents, as well as an F1 score computed only on discontinuous constituents (Disc.\ F1).
Following standard practice, we used the evaluation parameters included in discodop release (\texttt{proper.prm}).
These parameters ignore punctuation and root symbols.

\subsection{Optimization and Hyperparameters}
\label{sec:hyperparameters}

We optimize the loss with the Averaged Stochastic Gradient Descent algorithm
\cite{doi:10.1137/0330046,bottou-2010}
using the following dimensions for embeddings and hidden layers:
\begin{itemize}[noitemsep,topsep=0pt]
    \item Feedforward network: 2 layers of 128 units with rectifiers as activation function;
    \item The character bi-LSTM has 1 layer, with states of size 32 (in each direction);
    \item The sentence bi-LSTM has 2 layers, with states of size 128 (in each direction);
    \item Character embedding size: 32;
    \item Word embedding size: 32;
\end{itemize}
We tune the learning rate ($\{0.01, 0.02\}$) and the number of
iterations ($\{4, 8, 12, \dots, 28, 30\}$) on the development sets of each corpus.
All parameters, including embeddings, are randomly initialized.
We use no pretrained word embeddings nor any other external data.\footnote{We
leave to future work the investigation of the effect
of pretrained word embeddings and semi-supervised learning methods,
such as tritraining, that have been shown to be effective 
in recent work on projective constituency parsing
\cite{choe-charniak:2016:EMNLP2016,kitaev-klein:2018:Long}.}

Finally, following the method of \newcite{TACL885} to handle unknown words,
each time we sample a sentence from the training set, we stochastically replace each
word by an UNKNOWN pseudoword with a probability
$p_w = \left(\frac{\alpha}{\#\{w\} + \alpha} \right)$, where $\#\{w\}$ is the raw number
of occurrences of $w$ in the training set and $\alpha$ is a hyperparameter
set to $0.8375$, as suggested by \newcite{cross-huang:2016:EMNLP2016}.

\subsection{Runtime efficiency}
\label{sec:runtime}

For each experiment, we performed both training and parsing on a single CPU core.
Training a single model on the Tiger corpus (i.e.\ the largest training corpus)
took approximately a week.
Parsing the 5000 sentences of the development section of the
Tiger corpus takes 53 seconds (1454 tokens per second)
for the \textsc{ml-gap} model and 40 seconds (1934 tokens per second)
for the \textsc{sr-gap-unlex} model,
excluding model initialization and input-output times (Table~\ref{tab:runtime}).

Although indicative, these runtimes compare well
to other neural discontinuous parsers, e.g.\ \newcite{corro-leroux-lacroix:2017:EMNLP2017},
or to transition-based parsers using a linear classifier 
\cite{maier:2015:ACL-IJCNLP,coavoux-crabbe:2017:EACLlong}.

\begin{table}
\resizebox{\columnwidth}{!}{
    \begin{tabular}{lrrrr}
        \toprule
               Model                         & \multicolumn{2}{c}{Tiger} & \multicolumn{2}{c}{DPTB} \\
                                             & tok/s  & sent/s & tok/s  & sent/s \\
        \midrule
        This work, \textsc{ml-gap},       \textsc{base} & 1454 & 95   & 1450 & 61 \\
        This work, \textsc{sr-gap-unlex}, \textsc{base} & 1934 & 126  & 1887 & 80 \\
        \midrule
        \newcite{maier:2015:ACL-IJCNLP}, beam=8 & & 80 & & \\
        \newcite{coavoux-crabbe:2017:EACLlong}, beam=4 & 4700 & 260 & & \\
        \newcite{corro-leroux-lacroix:2017:EMNLP2017} &  & & & $\approx$7.3 \\
        \bottomrule
    \end{tabular}
}
\caption{Running times on development sets of the Tiger and the DPTB,
reported in tokens per second (tok/s) and sentences per second (sent/s).
Runtimes are only indicative, they are not comparable
with those reported by other authors,
since they use different hardware.}
\label{tab:runtime}
\end{table}

\subsection{Results}
\label{sec:results}

First, we compare the results of our proposed
models on the development sets, focusing on the effect of lexicalization
(Section~\ref{sec:result-lex}).
Then, we present morphological analysis results (Section~\ref{sec:morpho-comparisons}).
Finally, we compare our best model to other published results on the test sets (Section~\ref{sec:external-comparisons}).

\subsubsection{Effect of Lexicalization}
\label{sec:result-lex}

\paragraph{Lexicalized vs unlexicalized models}
We first compare the unlexicalized \textsc{ml-gap} system
with the \textsc{ml-gap-lex} system (Table~\ref{unlex:tab:parse-dev}).
The former consistently obtains higher results.
The F-score difference is small on English (0.1 to 0.3)
but substantial on the German treebanks (more than 1.0 absolute point)
and in general on discontinuous constituents (Disc.\ F).

In order to assess the robustness of the advantage of unlexicalized
models, we also compare our implementation of \textsc{sr-gap}
\cite{coavoux-crabbe:2017:EACLlong}\footnote{This is not
the same model as \newcite{coavoux-crabbe:2017:EACLlong}
since our experiments use the statistical
model presented in Section~\ref{sec:architecture},
with joint morphological analysis,
whereas they use a structured perceptron
and require a POS-tagged input.}
with an unlexicalized variant (\textsc{sr-gap-unlex})
that uses a single type of reduction
(\textsc{reduce}) instead of the traditional \textsc{reduce-right}
and \textsc{reduce-left} actions.
This second comparison exhibits the same pattern in favour
of unlexicalized models.

These results suggest that lexicalization is not necessary to achieve
very strong discontinuous parsing results.
A possible interpretation is that the bi-LSTM transducer
may implicitly learn latent lexicalization,
as suggested by \newcite{kuncoro-EtAl:2017:EACLlong},
which is consistent with recent analyses of other types
of syntactic information captured by LSTMs in parsing models
\cite{gaddy-stern-klein:2018:N18-1}
or language models \cite{Q16-1037}.

\paragraph{Effect of Lexical Features}
For lexicalized models, information about the head
of constituents (\textsc{+lex}) have a mixed effect
and brings an improvement in only half the cases.
It is even slightly detrimental on
English (\textsc{ml-gap-lex}).

\paragraph{Controlling for the oracle choice}
The advantage of unlexicalized systems could be due to the properties
of its eager oracle, in particular its higher incrementality
(see Section~\ref{sec:analysis} for an analysis).
In order to isolate the effect of the oracle,
we trained \textsc{ml-gap} with the head-driven oracle,
i.e.\ the oracle used by the \textsc{ml-gap-lex} system.
We observe a small drop in F-measure on English (-0.1) and on the Tiger corpus
(-0.4) but no effect on the Negra corpus.
However, the resulting parser still outperforms \textsc{ml-gap-lex},
with the exception of English.
These results suggest that the oracle choice definitely plays a
role in the advantage of \textsc{ml-gap} over \textsc{ml-gap-lex},
but is not sufficient to explain the performance
difference.

\paragraph{Discussion}
Overall, our experiments provide empirical arguments in favour of
unlexicalized discontinuous parsing systems.
Unlexicalized systems are arguably simpler than their lexicalized
counterparts -- since they have no directional (left or right) actions --
and obtain better results.
We further hypothesize that derivations produced by the eager
oracle, that cannot be used by lexicalized systems, 
are easier to learn.
We provide a quantitative and comparative analysis of derivations
from both transition systems in Section~\ref{sec:analysis}.

\begin{table}[t]
    \begin{center}
    \resizebox{\columnwidth}{!}{
        \begin{tabular}{llccr}
            \toprule
            Corpus&Attribute & Acc. & F1 & Cov. \\
            \midrule
    PTB &     POS            & 97.2 & - & 100 \\
            \midrule
    Negra&   POS            & 98.1 & - & 100 \\
            \midrule
    Tiger&  POS            & 98.4 & - & 100 \\
    (ours)& Complete match & 92.9 & - & 100 \\
            \cmidrule(r){2-2}
            &Case & 96.9 & 96.9 & 48.2 \\
            &Degree & 99.7 & 98.0 & 7.5 \\
            &Gender & 96.9 & 96.8 & 47.7 \\
            &Mood & 99.9 & 99.1 & 7.8 \\
            &Number & 98.4 & 98.7 & 57.8 \\
            &Person & 99.9 & 99.5 & 9.5 \\
            &Tense & 99.9 & 99.3 & 7.8 \\
            \midrule
    \multicolumn{5}{l}{Tiger \cite{bjorkelund-EtAl:2013:SPMRL,mueller-schmid-schutze:2013:EMNLP}}\\
            &POS             & 98.1 & & \\
            &Complete match  & 91.8 & & \\
            \bottomrule
        \end{tabular}
    }
    \end{center}
    \caption{Morphological analysis
    results on development sets.}
    \label{unlex:tab:results-morpho}
\end{table}

\begin{table*}
    \resizebox{\textwidth}{!}{
    \begin{tabular}{l cc cc cc}
        \toprule
        &\multicolumn{2}{c}{English (DPTB)} & \multicolumn{2}{c}{German (Tiger)} & \multicolumn{2}{c}{German (Negra)}  \\
        Model  & F & Disc. F & F & Disc. F & F & Disc. F \\
        \midrule
        \multicolumn{7}{c}{Predicted POS tags} \\
        \midrule
        Ours$^{*}$, \textsc{ml-gap} (\textsc{sr-gap-unlex} for Negra), \textsc{base} features & \textbf{91.0} 
        & \textbf{71.3} & \textbf{82.7} & \textbf{55.9} & \textbf{83.2} & \textbf{54.6} \\
        \midrule
        \newcite{stanojevic-garridoalhama:2017:EMNLP2017}$^{*}$, \textsc{swap}, stack/tree-LSTM & & & 77.0 & & & \\
        \newcite{coavoux-crabbe:2017:EACLlong}, \textsc{sr-gap}, perceptron                & &  & 79.3 & & &  \\
        \newcite{versley:2016:DiscoNLP}, pseudo-projective, chart-based                   & &  & 79.5 & & & \\
        \newcite{corro-leroux-lacroix:2017:EMNLP2017}$^{*}$, bi-LSTM, Maximum Spanning Arborescence       & 89.2    & & & & \\
        \newcite{vancranenburgh2016disc}, DOP, $\leq 40$                         & 87.0  & & & & 74.8\\
        \newcite{fernandezgonzalez-martins:2015:ACL-IJCNLP}, dependency-based             & &  & 77.3 & & & \\
        \newcite{gebhardt:2018:C18-1}, LCFRS with latent annotations   & &  & 75.1 & & & \\
        \midrule
        \multicolumn{7}{c}{Gold POS tags} \\
        \midrule
        \newcite{stanojevic-garridoalhama:2017:EMNLP2017}$^{*}$, \textsc{swap}, stack/tree-LSTM   & &  & 81.6 & & 82.9 \\
        \newcite{coavoux-crabbe:2017:EACLlong}, \textsc{sr-gap}, perceptron                       & &  & 81.6 & 49.2 & 82.2 & 50.0 \\
        \newcite{maier:2015:ACL-IJCNLP}, \textsc{swap}, perceptron   &   & & 74.7 & 18.8 &  77.0 & 19.8 \\
        \newcite{corro-leroux-lacroix:2017:EMNLP2017}$^{*}$ bi-LSTM, Maximum Spanning Arborescence  &  90.1 & & 81.6 & & \\
        \newcite{evang-kallmeyer:2011:IWPT}, PLCFRS, $< 25$ & 79$^\dagger$ & & & & \\
        \bottomrule
    \end{tabular}
    }
    \caption[Discontinuous parsing results (test).]{Discontinuous parsing
    results on the test sets.\\
    $^*$Neural scoring system.
    $^\dagger$Does not discount root symbols and punctuation.}
    \label{unlex:tab:parse-test}
\end{table*}

\subsubsection{Tagging and Morphological Analysis}
\label{sec:morpho-comparisons}

We report results for morphological analysis with the selected models
(\textsc{ml-gap} with \textsc{base} features for the Penn Treebank
and Tiger, \textsc{sr-gap-unlex} with \textsc{base} features for Negra)
in Table~\ref{unlex:tab:results-morpho}.
For each morphological attribute, we report an accuracy
score computed over every token.
However, most morphological attributes are only relevant
for specific part-of-speech tags.
For instance, \textsc{tense} is only a feature of verbs.
The accuracy metric is somewhat misleading, since the fact that the tagger
predicts correctly that a token does not have an attribute
is considered a correct answer.
Therefore, if only $5\%$ of tokens bore a specific morphological attribute,
a $95\%$ accuracy is a most-frequent baseline score.
For this reason, we also report a coverage metric (Cov.) that indicates
the proportion of tokens in the corpus that possess an attribute,
and an F1 measure.

The tagger achieves close to state-of-the-art results on all three corpora.
On the Tiger corpus, it slightly outperforms previous results published
by \newcite{bjorkelund-EtAl:2013:SPMRL} who used the \textsc{MarMoT} tagger
\cite{mueller-schmid-schutze:2013:EMNLP}.
Morphological attributes are also very well predicted, with F1 scores
above 98\%, except for case and gender.

\subsubsection{External Comparisons}
\label{sec:external-comparisons}

The two best-performing models on the development sets
are the \textsc{ml-gap} (DPTB, Tiger) and the \textsc{sr-gap-unlex}
(Negra) models with \textsc{base} features.
We report their results on the test sets
in Table~\ref{unlex:tab:parse-test}.
They are compared to other published results:
transition-based parsers using a \textsc{swap} action \cite{maier:2015:ACL-IJCNLP,stanojevic-garridoalhama:2017:EMNLP2017}
or a \textsc{gap} action \cite{coavoux-crabbe:2017:EACLlong},
the pseudo-projective parser of \newcite{versley:2016:DiscoNLP},
parsers based on non-projective dependency parsing \cite{fernandezgonzalez-martins:2015:ACL-IJCNLP,corro-leroux-lacroix:2017:EMNLP2017},
and finally chart parsers based on probabilistic
LCFRS \cite{evang-kallmeyer:2011:IWPT,gebhardt:2018:C18-1}
or data-oriented parsing \cite{vancranenburgh2016disc}.
Note that some of these publications report
results in a gold POS-tag scenario, a much easier
experimental setup that is not comparable to ours
(bottom part of the table).
In Table~\ref{unlex:tab:parse-test}, we also indicate models that use a neural scoring 
system with a `$^*$'.

Our models obtain state-of-the-art results
and outperform every other system, including the LSTM-based parser of
\newcite{stanojevic-garridoalhama:2017:EMNLP2017}
that uses a \textsc{swap} action to predict discontinuities.
This observation confirms in another setting the results of
\newcite{coavoux-crabbe:2017:EACLlong},
namely that \textsc{gap} transition systems
have more desirable properties
than \textsc{swap} transition systems.

\section{Model Analysis}
\label{sec:analysis}

In this section, we investigate empirical properties of
the transition systems evaluated in the previous section.
A key difference between lexicalized and unlexicalized
systems is that the latter are arguably simpler: they do not
have to assign heads to new constituents.
As a result, they need fewer types of distinct transitions,
and they have simpler decisions to make.
Furthermore, they do not run the risk of error propagation
from wrong head assignments.

We argue that an important consequence of the
simplicity of unlexicalized systems is that
their derivations are easier to learn.
In particular, \textsc{ml-gap} derivations
have a better incrementality than those of \textsc{ml-gap-lex}
(Section~\ref{sec:incrementality})
and are more economical in terms of number of \textsc{gap}
actions needed to derive discontinuous trees (Section~\ref{sec-gap-count}).

\subsection{Incrementality}
\label{sec:incrementality}

We adopt the definition of incrementality of \newcite{nivre:2004:IncrementalParsing}:
an incremental algorithm minimizes the number of connected components in the
stack during parsing.
An unlexicalized system can construct a new constituent by incorporating
each new component immediately whereas a lexicalized system waits until
it has shifted the head of a constituent before starting building the constituent.
For example, to construct the following head-final NP,
\begin{center}
    \footnotesize
    \begin{forest}
        [{NP[actor]} [An] [excellent] [environmental] [actor]]
    \end{forest}\\
\end{center}
a lexicalized system must shift every token before starting reductions
in order to be able to predict the dependency arcs between the head \textit{actor} and its three
dependents.\footnote{\textsc{sh(ift)}, \textsc{sh}, \textsc{sh}, \textsc{sh}, \textsc{sh}, \textsc{m(erge)-r(ight)}, \textsc{m-r}, \textsc{m-r}, \textsc{m-r}, \textsc{label-NP}. (\textsc{no-label} actions are omitted.)}
In contrast, an unlexicalized system can construct partial structures as soon
as there are two elements with the same parent node in the
stack.\footnote{\textsc{sh}, \textsc{sh}, \textsc{m(erge)}, \textsc{sh}, \textsc{m}, \textsc{sh}, \textsc{m}, \textsc{sh}, \textsc{m}, \textsc{label-NP}.}

We report the average number of connected components in the stack during a derivation
computed by an oracle for each transition system in Table~\ref{unlex:tab:incrementality}.
The unlexicalized transition system \textsc{ml-gap} has a better incrementality.
On average, it maintains a smaller stack. This is an advantage
since parsing decisions rely on information extracted from the stack
and smaller localized stacks are easier to represent.

\begin{table}
    \begin{center}
        \begin{tabular}{l rr}
        \toprule
        \multicolumn{3}{c}{Average length of stack (S+D)} \\
        Corpus         & \textsc{ml-gap-lex} & \textsc{ml-gap} \\
        \midrule
        English (DPTB)       & 5.62 & 4.86 \\
        German (Negra)       & 3.69 & 2.88 \\
        German (Tiger)       & 3.56 & 2.98 \\
        \bottomrule
        \end{tabular}
    \end{center}
    \caption{
    Incrementality
    measured by the average size of the stack during derivations.
    The average is calculated across all configurations (not across all sentences).}
    \label{unlex:tab:incrementality}
\end{table}

\begin{table*}
    \begin{center}
    \begin{tabular}{llrrr}
        \toprule
        &    &      \textsc{ml-gap-lex}      &  \textsc{ml-gap} \\
        \midrule
        English (DPTB) &Max number of consecutive \textsc{gaps}                & 9               & 8 \\
        &Average number of consecutive \textsc{gaps}         & 1.78            & 1.34 \\
        &Total number of \textsc{gaps}                      & 33,341           & 18,421 \\
        \midrule
        German (Tiger) &Max number of consecutive \textsc{gaps}               & 10              & 5 \\
        &Average number of consecutive \textsc{gaps}         & 1.40            & 1.12 \\
        &Total number of \textsc{gaps}                      & 40,905           & 25,852 \\
        \midrule
        German (Negra) & Max number of consecutive \textsc{gaps}               & 11              & 5 \\
        & Average number of consecutive \textsc{gaps}         & 1.47            & 1.11 \\
        & Total number of \textsc{gaps}                      & 20,149           & 11,181 \\
        \bottomrule
    \end{tabular}
    \end{center}
    \caption{\textsc{Gap} action statistics in training sets.}
    \label{unlex:tab:disco-stats}
\end{table*}

\subsection{Number of \textsc{gap} Actions}
\label{sec-gap-count}

The \textsc{gap} actions are supposedly the most difficult to predict,
because they involve long distance information.
They also increase the length of a derivation and make
the parser more prone to error propagation.
We expect that a transition system that is able to predict a discontinuous tree
more efficiently, in terms of number of \textsc{gap} actions, to be a better choice.

We report in Table~\ref{unlex:tab:disco-stats} the number of
\textsc{gap} actions necessary to derive the discontinuous trees
for several corpora and for several transition systems (using oracles).
We also report the average and maximum number of consecutive \textsc{gap} actions in each case.
For English and German, the unlexicalized transition system \textsc{ml-gap}
needs much fewer \textsc{gap} actions to derive discontinuous trees
(approximately 45\% fewer).
The average number of consecutive \textsc{gap} actions is also smaller (as well
as the maximum for German corpora).
In average, the elements in the stack ($S$) that need to combine
with the top of $D$ are closer to the top of $S$ with the
\textsc{ml-gap} transition system than with lexicalized systems.
This observation is not surprising, since \textsc{ml-gap} can start constructing
constituents before having access to their lexical head,
it can construct larger structures before having to \textsc{gap} them.

\section{Error Analysis}

In this section, we provide an error analysis focused on the
predictions of the \textsc{ml-gap} model on the
discontinuous constituents of the discontinuous PTB.
It is aimed at understanding which types of long distance dependencies
are easy or hard to predict and providing insights for future work.

\subsection{Methodology}

We manually compared the gold and predicted trees from the development set
that contained at least one discontinuous constituent.

Out of 278 sentences in the development set containing a discontinuity (excluding those in which
the discontinuity is only due to punctuation attachment),
165 were exact matches for discontinuous constituents and 113
contained at least one error.
Following \newcite{evang2011parsing}, we classified errors according
to the phenomena producing a discontinuity. We used the following typology,\footnote{These categories cover all cases in the development set.}
illustrated by examples where the main discontinuous constituent is highlighted in bold:
\begin{itemize}[noitemsep,topsep=0pt]
    \item \textit{Wh}-extractions: \textit{\textbf{What} should I \textbf{do}?}
    \item Fronted quotations: \textit{\textbf{``Absolutely''}, he \textbf{said}.}
    \item Extraposed dependent: \textit{In April 1987, \textbf{evidence} surfaced \textbf{that commissions were paid.}}
    \item Circumpositioned quotations: \textit{\textbf{In general}, they say, \textbf{avoid takeover stocks.}}
    \item It-extrapositions: \textit{``\textbf{It}'s better \textbf{to wait}.''}
    \item Subject-verb inversion: \textit{\textbf{Said} the spokeswoman: \textbf{``The whole structure has changed.''}}
\end{itemize}
For each phenomenon occurrence, we manually classified the output
of the parser in one the following categories
(i) perfect match (ii) partial match (iii) false negative.
Partial matches are cases where the parser identified
the phenomenon involved but made a mistake regarding
the labelling of a discontinuous constituent (e.g.\ S instead of SBAR) or
its scope.
The latter case includes e.g.\ occurrences where the parser
found an extraction, but failed to find the correct extraction
site.
Finally, we also report false positives for each phenomenon.

\subsection{Results}

First of all, the parser tends to be conservative when predicting discontinuities,
there are in general few false positives.
The 72.0 discontinuous F1 (Table~\ref{unlex:tab:parse-dev})
indeed decomposes in a precision of 78.4 and a recall of 66.6.
This does not seem to be a property of our parser, as other
authors also report systematically higher precisions than recalls \cite{maier:2015:ACL-IJCNLP,stanojevic-garridoalhama:2017:EMNLP2017}.
Instead, the scarcity of discontinuities in the data might be a determining
factor: only 20\%~of sentences in the Discontinuous Penn Treebank contain
at least one discontinuity and 30\%~of sentences in the Negra and Tiger corpus.

\begin{table}[t]
    \resizebox{\columnwidth}{!}{
    \begin{tabular}{lr|rrr|r}
        \toprule
        Phenomenon  & G & PfM & PaM & FN & FP \\
        \midrule
        \textit{Wh}-extractions & 122   & 87 & 19 & 16 & 8 \\
                                & 100\% & 71.3 & 15.6 & 13.1 & NA \\
        \midrule
        Fronted quotations  & 81 & 77 & 3 & 1 & 0   \\
                            & 100\% & 95.1 & 3.7 & 1.2 & NA \\
        \midrule
        Extrapositions      & 44 & 10 & 1   & 33 & 3 \\
                            & 100\% & 22.7 & 2.3 & 75 & NA\\
        \midrule
        Circumpositioned quotations & 22 & 11 & 10 & 1 & 3 \\
                    & 100\% & 50 & 45.4 & 4.5 & NA \\
        \midrule
        \textit{It}-extrapositions & 16 & 6 & 2 & 8 & 2 \\
                            & 100\% & 37.5 & 12.5 & 50 & NA \\
        \midrule
        Subject-verb inversion & 5 & 4 & 0 & 1 & 1 \\
                        & 100\% & 80 & 0 & 20 & NA \\
        \bottomrule
    \end{tabular}}

    \caption{Evaluation statistics per phenomenon.
    G: gold occurrences, PfM: perfect match, PaM: partial match, FN: false negatives, FP: false positives.}
    \label{unlex:tab:error-analysis-synthesis}
\end{table}

Analysis results are presented in Table \ref{unlex:tab:error-analysis-synthesis}.
For \textit{wh}-extractions, there are two main causes of errors.
The first one is an ambiguity on the extraction site.
For example, in the relative clause \textit{which many clients didn't know about},
instead of predicting a discontinuous PP, where \textit{which} is the complement of \textit{about},
the parser attached \textit{which} as a complement of \textit{know}.
Another source of error (both for false positives and false negatives) is the ambiguity of \textit{that}-clauses,
that can be either completive clauses\footnote{(NP the consensus \dots (SBAR that the Namibian guerrillas were above
all else the victims of suppression by neighboring South Africa.))}
or relative clauses.\footnote{(NP the place (SBAR \textbf{that} world opinion has been celebrating \textbf{over}))}

Phenomena related to quotations are rather well identified
probably due to the fact that they are frequent in newspaper data and exhibit
regular patterns (quotation marks, speech verbs).
However, a difficulty in identifying circumpositioned quotations arises
when there are no quotation marks, to determine what the precise scope of the
quotation is.

Finally, the hardest types of discontinuity for the parser
are extrapositions. Contrary to previously discussed phenomena, there
is usually no lexical trigger (\textit{wh}-word, speech verb)
that makes these discontinuities easy to spot.
Most cases involve modifier attachment ambiguities, which are known to be
hard to solve \cite{kummerfeld-EtAl:2012:EMNLP-CoNLL} and often require some world knowledge.

\section{Conclusion}

We have introduced an unlexicalized transition-based
discontinuous constituency parsing
model.\footnote{The source code of the parser is released with pretrained models at \texttt{\url{https://github.com/mcoavoux/mtg_TACL}}.}
We have compared it, in identical experimental settings,
to its lexicalized counterpart
in order to provide insights on the effect of
lexicalization, as a parser design choice.

We found that lexicalization is not necessary
to achieve very high parsing results in discontinuous constituency parsing,
a result consistent with previous studies on lexicalization
in projective constituency parsing \cite{klein-manning:2003:ACL,cross-huang:2016:EMNLP2016}.
A study of empirical properties of our transition
systems suggested explanations
for the performance difference, by showing that
the unlexicalized system produces shorter derivations and has
a better incrementality.
Finally, we presented a qualitative analysis of our parser's
errors on discontinuous constituents.

\section*{Acknowledgments}
We thank Kilian Evang and Laura Kallmeyer for providing us with the discontinuous Penn Treebank.
We thank Caio Corro, Sacha Beniamine, TACL reviewers and
action editor Stephen Clark for feedback that helped improve the paper.
Our implementation makes use of the Eigen C++ library \cite{eigenweb}, \texttt{treetools}\footnote{\url{https://github.com/wmaier/treetools}} and \texttt{discodop}.\footnote{\url{https://github.com/andreasvc/disco-dop}}
MC and SC gratefully acknowledge the
support of Huawei Technologies.

\bibliographystyle{acl_natbib}
\bibliography{tacl2018}

\end{document}